\documentclass{article}
\usepackage{spconf,amsmath,graphicx}
\usepackage{algorithm}
\usepackage{algorithmicx}
\usepackage{algpseudocode}
\usepackage{amsfonts,bm}
\usepackage{amsthm}
\usepackage{amssymb}
\usepackage{footnote}
\usepackage{perpage}
\usepackage{url}
\usepackage{float}
\usepackage{stfloats}
\usepackage{mathrsfs}
\usepackage{booktabs}
\usepackage{multirow}
\renewcommand{\paragraph}[1]{\noindent\textbf{#1}\quad}
\newcommand{\tabincell}[2]{\begin{tabular}{@{}#1@{}}#2\end{tabular}}

\def\X{{\mathbf X}}

\def\Z{{\mathbf Z}}
\def\C{{\mathbf C}}

\usepackage{paralist}

\def\w{wav2vec2.0}

\title{Applying Wav2vec2.0 to Speech Recognition in Various \\ Low-resource Languages}
\name{Cheng Yi$^{1,2}$, Jianzhong Wang$^{3}$, Ning Cheng$^{3}$, Shiyu Zhou$^1$, Bo Xu$^{1,2}$}
\address{
    $^1$Institute of Automation, Chinese Academy of Sciences, China\\
    $^2$University of Chinese Academy of Sciences, China\\
    $^3$Ping An Technology (Shenzhen) Co., Ltd.\\
    \small{\{yicheng2016,zhoushiyu2013,xubo\}@ia.ac.cn}, jzwang@188.com, chengning211@pingan.com.cn}

\begin{document}
\maketitle

\begin{abstract}
   There are several domains that own corresponding widely used feature extractors, such as ResNet, BERT, and GPT-x. These models are usually pre-trained on large amounts of unlabeled data by self-supervision and can be effectively applied to downstream tasks. In the speech domain, wav2vec2.0 starts to show its powerful representation ability and feasibility of ultra-low resource speech recognition on the Librispeech corpus, which belongs to the audiobook domain. However, wav2vec2.0 has not been examined on real spoken scenarios and languages other than English.
   To verify its universality over languages, we apply pre-trained models to solve low-resource speech recognition tasks in various spoken languages. We achieve more than 20\% relative improvements in six languages compared with previous work. Among these languages, English achieves a gain of 52.4\%. Moreover, using coarse-grained modeling units, such as subword or character, achieves better results than fine-grained modeling units, such as phone or letter.
\end{abstract}

\section{Introduction}
Traditional speech recognition frameworks decompose the whole automatic speech recognition (ASR) task into acoustic, pronunciation, and language modeling \cite{povey2011kaldi}. 
End-to-end modeling has avoided the complex modeling process and human-designed lexicon at the cost of large amounts of labeled data.
It integrates the acoustic and the linguistic models into a single architecture that is jointly optimized. Emerging end-to-end models have outperformed traditional models in the ASR task over many datasets.

However, on many low-resource scenarios, the demand for labeled data for the end-to-end modeling cannot be satisfied. Especially, ASR models require much more data to learn the invariability of speech content \cite{zhou2017multilingual,zhou2018multilingual} on noisy or spoken scenarios.
% low-resource & pre-train
One mainstream research paradigm for the end-to-end modeling is to pre-train the model partially in supervised \cite{inaguma2019transfer} or self-supervised way \cite{schneider2019wav2vec,oord2018representation,devlin2019bert}. By pre-training, models can learn general representation and fit downstream tasks from a good starting point  \cite{mikolov2013distributed,devlin2019bert}, greatly reducing the parameter searching space and relieving over-fitting.
% For low-resource ASR tasks that usually contain less than 20 hours of transcribed speech, effective acoustic pre-training can vastly reduce the difficulty of learning to process the input speech.

% supervised Pre-train
Supervised pre-training methods include task transfer \cite{zhou2018multilingual} and multi-task learning \cite{Lugosch2019}. They converge fast while the learned representation is specialized towards solving a single task. The similarity of data distribution between the warm-up task and target task has a huge impact on the effectiveness of pre-training.
% self-supervised pre-train
In contrast, self-supervised pre-training can push models to explore robust and general representation at the expense of a large amount of computation.
% auto-regressive pre-train
% BERT \cite{devlin2018bert} is the first model which successfully achieves significant improvements in various text-process tasks. To process the audio signal inputs, \cite{jiang2019improving} imitates BERT to mask input feature sequence and predict the missing parts of the input. \cite{chorowski2019unsupervised} uses auto-encoding structure learn discrete latent.
% However, this auto-regressive self-training needs to recover the original high-dimension input in detail from the representation, which is unnecessary for most downstream speech tasks.
% Contrastive pre-train
Contrastive Predictive Coding (CPC) is proposed as a self-supervised training criterion to learn the representations from the input source by predicting the future in its latent space \cite{oord2018representation}. 
Wav2vec introduces CPC on the ASR task and achieves a better speech feature extractor than human-designed ones \cite{schneider2019wav2vec}.
Vq-wav2vec \cite{baevski2019vqwav2vec} quantizes the audio representation (continuous vectors) into elements which belongs to a fixed dictionary. Further standard BERT pre-training is applied to the discrete sequences. 
Wav2vec2.0 fuses the BERT masked sequence modeling and discrete CPC training into a whole model \cite{baevski2020wav2vec}. It outperforms previous work on the 100-hour subset while using 100 times less labeled data. By using just 10 minutes of transcribed speech, it achieves 5.7/10.1 WER on the noisy/clean test sets of Librispeech \cite{panayotov2015librispeech}.

\begin{figure*}[t]
    \centering
    \includegraphics[width=0.75\linewidth]{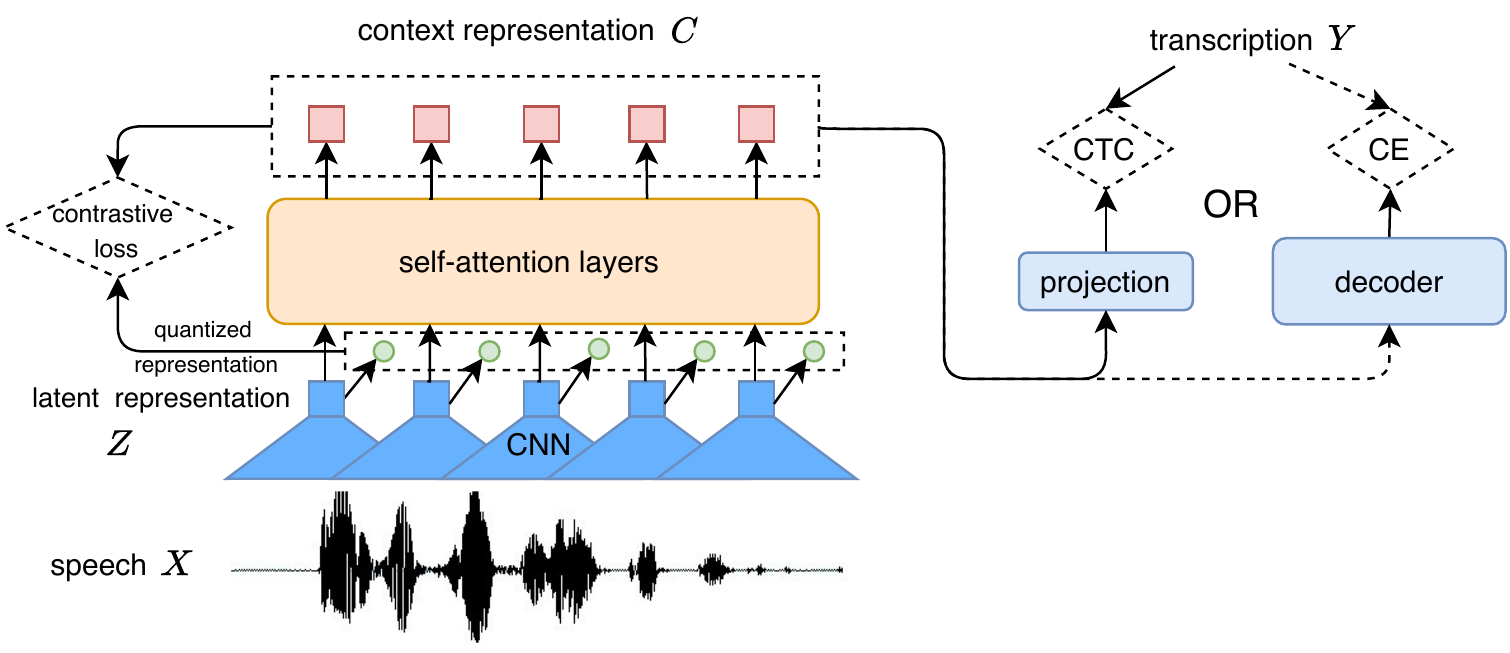}
    \caption{Left: the structure of \w\ and corresponding self-training criterion. It contains a stack of convolution layers and self-attention layers; Right: two decoding branches that apply wav2vec2.0 to ASR tasks with additional projection or decoder, which is trained with CTC or cross-entropy loss respectively.}
    \label{fig:wav2vec}
\end{figure*}

This work focuses on applying \w\ pretrained using English speech to solve low-resource ASR tasks in six languages. These corpora are recorded in real spoken scenarios, which is quite different from the speech used for pre-training.
The key contributions of our work can be concluded as follows:
\begin{inparaenum}[\it 1)]
    \item We show that \w\ can work well on the real low-resource ASR task in various spoken languages with a low sampling rate (8k). 
    \item We demonstrate that \w\ can adapt to coarse-grained modeling unit and generally achieve better performance free from pronunciation modeling. 
    \item We carry out sufficient experiments and analysis, and discover many interesting and important phenomena.
    % It is not straightforward since the model is pre-trained with frame-level criterion.
\end{inparaenum}

% This work focuses on applying \w\ to solve low-resource ASR tasks in six languages. These corpora are recorded in real spoken scenarios, which is quite different from the speech used for pre-training. The key contributions of our work include:
% \begin{itemize}
% \setlength\itemsep{0.1em}
% \item We show that \w\ can work well in various spoken languages with a low sampling rate (8k), considering it is only pre-trained on English audio signals sampled by 16k. We conjecture \w\ has learned basis acoustic units that can compose diverse languages. 
% \item We verify that models pre-trained with frame-level self-supervised learning can apply coarse modeling unit, performing better and free from pronunciation modeling. It shows that \w\ can dynamically merge the fine-grained knowledge into coarser ones to fit the target task.    
% \end{itemize}

\section{Methodology}

Firstly, we describe the structure of \w. Then we simply recall the self-supervised pre-training criterion proposed in \cite{baevski2020wav2vec}. Finally, two common ways are introduced to apply \w\ to solve ASR tasks.

\label{sec:model}
    \subsection{Wav2vec2.0 Structure}
    \label{ssec:wav2vec}
    
    As shown in Fig \ref{fig:wav2vec}, \w \ consists of multiple convolution layers and self-attention layers. This structure is widely used in recent end-to-end ASR models \cite{zhou2018multilingual,dong2020cif,li2019speechtransformer}. 
    % The difference is that \w\ directly processes the speech $\X$.
    Convolution layers down-sample speech $\X$ and generate more compressed latent representation $\Z$. Specifically, $\Z$ represents the raw audio signals $\X$ sampled by 16k with a stride of about 20ms and a receptive ﬁeld of 25ms. 
    Self-attention layers \cite{waswani2017attention} build contextualized representations $\C$ and capture high level content from input $\Z$. Their strong context dependency modeling ability empowers the model to make the right choices during the following contrastive training given the masked $\Z$. 
    
    \subsection{Self-training Criterion}
    \label{ssec:discriminate}
    \begin{figure}[h]
    \centering
    \includegraphics[width=1.0\linewidth]{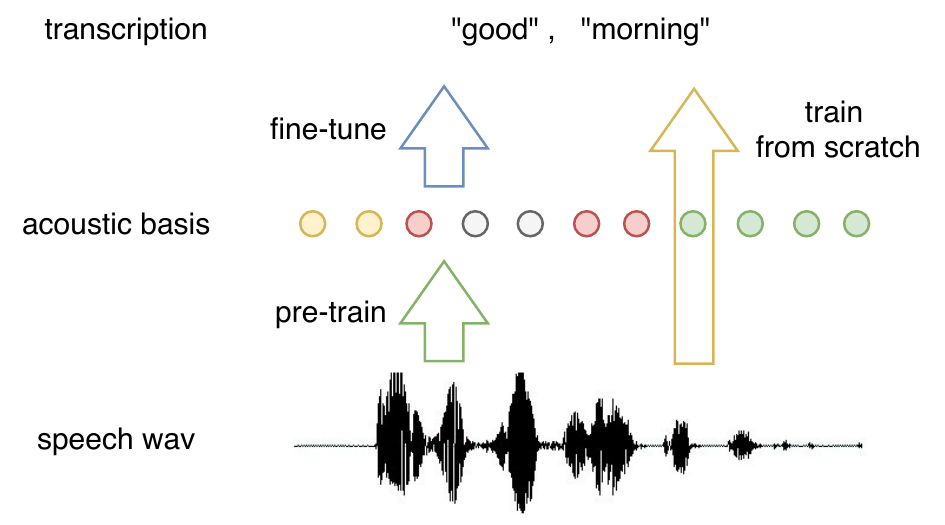}
    \caption{Demonstration of how \w \ pre-training works. The model is pushed to learn distinguishable fine-grained acoustic units during pre-training phase. They will merge into coarser-grained ones to fit the target ASR task during the fine-tuning phase.
    }
    \label{fig:pretrain}
    \end{figure}

    BERT is the first large-scale pre-trained model that successfully achieves significant improvements on various text processing tasks \cite{devlin2019bert}. To process the speech input, previous studies imitate BERT to mask the input feature sequence and predict the missing parts of the input \cite{jiang2019improving}, or use the auto-encoding structure to learn discrete latent representation \cite{chorowski2019unsupervised}.
    However, this auto-regressive self-training requires models to totally recover the original high-dimension input from the hidden representation. It is unreasonable and unnecessary for most downstream tasks. Wav2vec2.0 adopts CPC as the core training criterion, including contrastive training and discrete representation. 
    
    \paragraph{Contrastive Training} $\C$ is used for contrastive learning \cite{oord2018representation} conditioned on the masked $\Z$.
    Different from auto-regressive training, contrastive training only asks the model to distinguish the representation at masked time steps from other time steps. The change from a regressive task to a classification task leads to more effective self-training. 
    
    \paragraph{Discrete Representation} 
    % It is not an appropriate assumption to list all time steps but the current one as the wrong categories, which is actually solving a classification task with a huge number of categories (only single positive case within sequence length number of sampling all the time).
    Within the sequence length number of samples, there is only a single positive case. It is not an appropriate assumption to mark all time steps except for the current one as the negative categories, which is actually solving a classification task with a huge number of negetive samples.
    Quantization over $\Z$ shrinks the size of categories to a rather small one, relieving the sparseness of learning signals. 
    We conjecture that the quantization and contrastive training force the encoder to discover limited discrete acoustic units that are distinguishable from each other, as shown in Fig \ref{fig:pretrain}.
    % In practice, model chooses one entry from each codebook and concatenate as the identity of its category.

    % The self-supervised learning loss $\L$ contains constrastive loss $\Lm$, codebook diversity loss $\L_d$ and L2 penalty $\L_f$.

\begin{table*}[ht]
    \centering
    \caption{Performance on CALLHOME corpus with 6 languages. The criteria is WER(\%) on each test set (CER(\%) for MA and JA). }
    \setlength{\tabcolsep}{5mm}
    \vspace{4pt}
    \begin{tabular}{ l c c c c c c }
    \toprule
    \textbf{Models} & \textbf{AR} & \textbf{EN} & \textbf{MA} & \textbf{JA} & \textbf{GE} & \textbf{SP} \\
    \midrule
    mlstm-residual \cite{zhou2017multilingual} & 56.47 & 43.93 & 45.85 & 50.13 & 51.75 & 53.38  \\
    \midrule
     Speech-Transformer \cite{zhou2018multilingual} & 48.35 & 33.77 & 37.62 & 36.99 & 44.98 & 51.54  \\
    \bottomrule
    \tabincell{l}{wav2vec2.0-base \\ \quad  + ctc (letter/phone)\\ \quad  + LM decode} & 40.73 & 21.83 & 33.57 & 38.24 & 29.88 & 45.92 \\
    \midrule
    \tabincell{l}{wav2vec2.0-base \\ \quad + ctc (subword/char) } & 50.67 & 24.93 & 36.06 & 37.70 & 41.77 & 52.53 \\
    + LM decode & 40.63 & 21.31 & 32.34 & 36.28 & 33.83 & 45.50 \\
    \midrule
    \tabincell{l}{wav2vec2.0-large \\ \quad + ctc (letter/char)} & 42.44 & 17.65 & 28.75 & 28.69 & 40.27 & 47.36 \\
    + LM decode & \textbf{35.62} & \textbf{16.07} & \textbf{28.16} & \textbf{28.32} & \textbf{25.70} & \textbf{39.11} \\
    \bottomrule
    relative improvement & 26.3\% & \textbf{52.4
    }\% & 25.1\% & 23.4\% & 42.9\% & 24.1\% \\
    \bottomrule
    \end{tabular}
\label{tab:compare1}
\end{table*}
    
    \subsection{Adaptation to ASR Tasks}
    \label{ssec:asr}
    As shown in Fig \ref{fig:wav2vec}, \w\ is used as an encoder. By adding a projection layer, it can be trained with the CTC criterion and directly generate modeling units in a frame-synchronous way \cite{baevski2020wav2vec,graves2012sequence}. Alternatively, a self-attention decoder can further process the context representation and generate modeling units in a label-synchronous way \cite{li2019speechtransformer}. 
    
    % modeling unit and pronunciation modeling
    Through the analysis in Section \ref{ssec:discriminate}, the encoder could have discovered acoustic units with 20ms interval by self-training, which can fit various languages. 
    It is natural to choose fine-grained output units \cite{baevski2020wav2vec}, such as letters or phonemes. In that case, however, further pronunciation modeling is needed to generate the final transcription. By contrast, coarse-grained grapheme as the modeling unit can directly compose the sentences. So we also experiment with subword and character as modeling units besides the letter. 
    
    % language modeling
    External language models can be used as the prior bias during beam decoding on the output of the model, which is a sequence of probabilistic distribution over the modeling units. Both the N-gram model and the neural network can be used as a language model \cite{baevski2020wav2vec}. 
    
\section{Experiments}
\label{sec:experiments}
We use publicly released pre-trained \w\ models: wav2vec2.0-base and wav2vec2.0-large, where the base model is trained on 1,000 hours and the large model is trained on 50,000 hours \footnote{\url{https://dl.fbaipublicfiles.com/fairseq/wav2vec/}}. We inherit most of the training configurations in the public repository. 
% We only use a single GPU (TITAN Xp) for each experiment, which costs less than 15 hours. We train with Adam optimizer, warming up the learning rate for 8000 steps to a peak of $4\times10^{-5}$, holding 42000 steps and then exponential decay it.
We train our models on 1 NVIDIA TITAN XP GPU for less than 15 hours using Adam optimizer. And we warm up the learning rate for 8000 steps to a peak of $4\times10^{-5}$, hold for 42000 steps and then exponential decay it.
After fine-tuning, we choose \textit{wav2letter++} decoder to get LM-biased results \cite{pratap2018wav2letter++}.

     Our experiments are mainly conducted on CALLHOME corpus, including six languages: Mandarin (MA), English (EN), Japanese (JA), Arabic (AR), German (GE), and Spanish (SP). Each one is an 8k sampled telephone conversation dataset with about 15 hours of transcribed speech. Details are shown in Table \ref{tab:dataset}. 
     We choose different modeling units:
     \begin{inparaenum}[\it a)]
     \item MA and JA belong to character-based writing systems and we simply count the characters appeared in the training set. We also apply phoneme for acoustic modeling, according to the lexicons provided by the corpus. 
     \item For the other four languages, modeling units are determined by the BPE algorithm as in \cite{zhou2018multilingual}.
     \end{inparaenum}
     The HKUST corpus is a similar Mandarin speech dataset of 150 hours with accent \cite{liu2006hkust}. 8k wav files are up-sampled to 16k to keep consistent with the pretrained model.
     
     \begin{table}[h]
        \centering
        \caption{Statistics and modeling units of CALLHOME dataset.}
        \vspace{4pt}
        \begin{tabular}{ l c c c }
        \toprule
        \textbf{Lang} & \textbf{training utts} & \textbf{test utts} & \textbf{unit}\\
        \midrule
        AR & 20k & 2.9k & subword (4768) \\
        EN & 21k & 2.8k & subword (1627) \\
        MA & 24k & 3.0k & character (2972) \\
        JA & 27k & 3.3k & character (2010) \\
        GE & 20k & 5.2k & subword (1763) \\
        SP & 18k & 1.9k & subword (3860) \\
        \bottomrule
        \end{tabular}
        \label{tab:dataset}
    \end{table}
     
    \subsection{Comparison wth Various Languages}
    \label{ssec:exp1}
    We add an additional projection layer and ﬁne-tune the ASR model with CTC loss. 5-gram models are used during decoding where each one is trained on corresponding training transcriptions.
    
    As we can see in Table \ref{tab:compare1}, models based on wav2vec2.0 surpass previous work in all languages, where wav2vec2.0-large models further exceed base ones by a large margin. It reveals the large model has much stronger acoustic representation.
    It is worth noting that EN has a distinct 52.4\% relative improvement among six languages. GE is more similar to EN in acoustics, and it can achieve a 42.9\% relative improvement. We conclude that adding abundant speech data in target language to pre-train can benefit the performance of ASR.
    
    \subsection{Effect of Modeling Unit}
    \label{ssec:exp2}
    Relative results are listed in Table \ref{tab:compare1}.
    Among the six languages, MA and JA (non-Latin languages) cannot use fine-grained grapheme as modeling unit. They adopt character as the coarse-grained unit, while other languages (Latin languages) use subword \cite{sennrich2016neural}. We can see that, in most cases, the coarse-grained unit is slightly better than fine-grained one. Moreover, coarse-grained units can directly generate recognized results without pronunciation modeling. 
    
    % language bias 
    For models with letter as modeling unit, we set the LM weight to 0 to get the acoustic recognized results. External language modeling can always bring significant improvement, regardless of languages, model size, or modeling unit. 
    
    \subsection{Results of the Encoder-decoder Structure}
    
    Previous work has demonstrated the importance of language modeling for ASR task. Especially, CTC models are unable to utilize the partly-decoded results during inference \cite{graves2012sequence,yi2019ectc,sak2017recurrent}. In this section, we introduce \w\ into the encoder-decoder framework. 
    An additional self-attention decoder \cite{bahdanau2015neural} uses cross attention to connect the encoder \cite{waswani2017attention}, as shown in Fig \ref{fig:wav2vec}. We compare different numbers of decoder layers on MA and HKUST. 
    % \label{ssec:w2v_seq2seq}
        \begin{table}[h]
        \centering
        \caption{Applying \w\ to the encoder-decoder structure. The criterion is CER (\%).}
        \setlength{\tabcolsep}{5mm}
        \vspace{4pt}
        \begin{tabular}{ l l l l }
        \toprule
        \textbf{Additional modules} & \textbf{MA} & \textbf{HKUST} \\
        \midrule
        train set & 15h & 150h \\
        \bottomrule
        1 projection layer (ctc) & \textbf{36.06} & \textbf{23.67} \\
        1 decoder layer (ce) & 39.81 & 24.06 \\
        4 decoder layers (ce) & 54.82 & 25.73 \\
        \bottomrule
        \end{tabular}
        \label{tab:compare3}
    \end{table}
    
    The results are listed in Table \ref{tab:compare3}. We find that encoder-decoder structure cannot achieve better results in all conditions. It becomes worse with limited data. We conclude that the amount of transcriptions in the low-resource task is too sparse for the decoder to generalize well. 
    
    \subsection{Self-supervised Vs Supervised Pre-training}
    In this section, we compare the effectiveness of self-supervised and supervised pre-training according to the fine-tuned ASR performance on the MA corpus. 
    The self-supervised pre-training refers to the criterion introduced in Section \ref{ssec:discriminate}, while the supervised pre-training means transferring a model already converged on some labeled speech datasets \cite{zhou2018multilingual}. 
    All the experiments are operated on \w-small.
    We consider two corpora for supervised pre-training: train-clean (100h) in Librispeech and HKUST training set (150h). The former is a subset dataset which is used to self-supervised pre-train, and the latter is similar to the target CALLHOME-MA dataset. We follow the Kaldi recipe to get phoneme alignments on Librispeech\footnote{\url{https://github.com/kaldi-asr/kaldi/tree/master/egs/librispeech}}. So we compare supervised pre-training on three different labeled data: Librispeech with subword transcriptions (EN subword), Librispeech with phoneme transcriptions (EN phone), and HKUST with character transcriptions (HKUST char).
    
    \begin{table}[h]
        \centering
        \caption{Supervised pre-training vs self-supervised pre-training. The criteria is CER(\%) on CALLHOME-MA test set.}
        \vspace{4pt}
        \begin{tabular}{ l l c }
        \toprule
        \textbf{Pre-training} & \textbf{Pre-training data} & \textbf{CER} \\
        \midrule
        w/o pre-train & None & 79.34 \\
        \midrule
        \multirow{3}{*}{supervised} 
        & Libri 100h speech + subword & 63.12 \\
        & Libri 100h speech + phone & 50.76 \\
        & HKUST 150h speech + char &  37.62 \\
        \midrule
        self-supervised & Libri 1000h speech & \textbf{36.06}\\
        \bottomrule
        \end{tabular}
    \label{tab:compare4}
    \end{table}
    
    As shown in Table \ref{tab:compare4}, supervised pre-training on other languages is weaker than that on the target-similar dataset. 
    However, self-supervised pre-training can make better use of other languages with rich audio data. It can even surpass the target-similar supervised pre-training, pointing a way to solve the low-resource tasks where the speech data is also limited.

    % \label{ssec:w2v_seq2seq}
    
\section{Conclusions}
\label{sec:conclusion}
In this work, we apply the pre-trained \w \ model to solve the low-resource ASR task. We verify that the encoder can still perform well on the ASR tasks in various languages. Using coarse-grained modeling units can achieve slightly better results. Applying self-supervised training can take full advantage of audio data in not only target language but also other languages.
We conjecture \w\ has learned basis acoustic units that can compose diverse languages. 
% It shows that \w\ can dynamically merge the fine-grained knowledge into coarser-grained ones to fit the target task.   
And our analysis shows that \w \ can dynamically merge the fine-grained presentation into coarser-grained presentation to fit the target task. 

\vfill\pagebreak
\bibliographystyle{IEEEtran}
\bibliography{mybib}

\end{document}